\documentclass[11pt,a4paper]{article}
\usepackage[hyperref]{emnlp2020}
\usepackage{times}
\usepackage{latexsym}
\usepackage{array, multirow}
\usepackage{tabularx, lipsum}
\usepackage{float}
\usepackage{soul}
\usepackage{graphicx}

\def\thick{\noalign{\hrule height 1pt}}
\newcolumntype{s}{>{\hsize=.45\hsize}X}

\usepackage{url}
\usepackage{microtype}
\usepackage{breakurl}
\usepackage{bm}
\usepackage{amsmath}

\aclfinalcopy 
\def\aclpaperid{3278} 

\title{Assessing Phrasal Representation and Composition in Transformers}

\author{Lang Yu \\
  Deptartment of Computer Science \\
  University of Chicago \\
  \texttt{langyu@uchicago.edu} \\\And
  Allyson Ettinger \\
  Department of Linguistics \\
  University of Chicago \\
  \texttt{aettinger@uchicago.edu} \\}

\date{}

\begin{document}
\maketitle
\begin{abstract}

Deep transformer models have pushed performance on NLP tasks to new limits, suggesting sophisticated treatment of complex linguistic inputs, such as phrases. However, we have limited understanding of how these models handle representation of phrases, and whether this reflects sophisticated composition of phrase meaning like that done by humans. In this paper, we present systematic analysis of phrasal representations in state-of-the-art pre-trained transformers. We use tests leveraging human judgments of phrase similarity and meaning shift, and compare results before and after control of word overlap, to tease apart lexical effects versus composition effects. We find that phrase representation in these models relies heavily on word content, with little evidence of nuanced composition. We also identify variations in phrase representation quality across models, layers, and representation types, and make corresponding recommendations for usage of representations from these models.
\end{abstract}

\section{Introduction}

A fundamental component of language understanding is the capacity to combine meaning units into larger units---a phenomenon known as \emph{composition}---and to do so in a way that reflects the nuances of meaning as understood by humans. Transformers \cite{vaswani2017attention} have shown impressive performance in NLP, particularly transformers using pre-training, like BERT \cite{devlin2019bert} and GPT \cite{radford2018improving, radford2019language}, suggesting that these models may be succeeding at composition of complex meanings. However, because transformers (like other contextual embedding models) typically maintain representations for every token, it is unclear how and at what points they might be combining word meanings into phrase meanings. This contrasts with models that incorporate explicit phrasal composition into their architecture, e.g. RNNG \cite{dyer2016recurrent, kim2019unsupervised}, recursive models for semantic composition \cite{socher2013recursive}, or transformers with attention-based composition modules \cite{yin2020sentibert}.

In this paper we take steps to clarify the nature of phrasal representation in transformers. We focus on representation of two-word phrases, and we prioritize identifying and teasing apart two important but distinct notions: how faithfully the models are representing information about the \textit{words} that make up the phrase, and how faithfully the models are representing the nuances of the \emph{composed phrase meaning} itself, over and above a simple account of the component words. To do this, we begin with existing methods for testing how well representations align with human judgments of meaning similarity: similarity correlations and paraphrase classification. We then introduce controlled variants of these datasets, removing cues of word overlap, in order to distinguish effects of word content from effects of more sophisticated composition. We complement these phrase similarity analyses with classic sense selection tests of phrasal composition~\cite{kintsch2001predication}. 

We apply these tests for systematic analysis of several state-of-the-art transformers: BERT \cite{devlin2019bert}, RoBERTa \cite{liu2019roberta}, DistilBERT \cite{sanh2019distilbert}, XLNet \cite{yang2019xlnet} and XLM-RoBERTa \cite{conneau2019unsupervised}. We run the tests in layerwise fashion, to establish the evolution of phrase information as layers progress, and we test various tokens and token combinations as phrase representations. We find that when word overlap is not controlled, models show strong correspondence with human judgments, with noteworthy patterns of variation across models, layers, and representation types. However, we find that correspondence drops substantially once word overlap is controlled, suggesting that although these transformers contain faithful representations of the lexical content of phrases, there is little evidence that these representations capture sophisticated details of meaning composition beyond word content. Based on the observed representation patterns, we make recommendations for selection of representations from these models. All code and controlled datesets are made available for replication and application to additional models.\footnote{Datasets and code available at \url{https://github.com/yulang/phrasal-composition-in-transformers}}

\section{Related work}

This paper contributes to a growing body of work on analysis of neural network models. Much work has studied recurrent neural network language models \cite{linzen2016assessing, wilcox2018rnn, chowdhury2018rnn, gulordava2019colorless, futrell2019neural}
and sentence encoders \cite{adi2016fine, conneau2018you, ettinger2016probing}. Our work builds in particular on analysis of information encoded in contextualized token representations \cite{bacon2019does, tenney2019you,peters2018dissecting,hewitt2019structural,klafka2020spying} and in different layers of transformers \cite{tenney2019bert, jawahar2019does}. The BERT model has been a particular focus of analysis work since its introduction. Previous work has focused on analyzing the attention mechanism \cite{vig2019analyzing, clark2019does}, parameters \cite{roberts2020much, radford2019language, raffel2020exploring} and embeddings \cite{shwartz2019still, liu2019linguistic}. We build on this work with a particular, controlled focus on the evolution of phrasal representation in a variety of state-of-the-art transformers.

Composition has been a topic of frequent interest when examining neural networks and their representations. One common practice relies on analysis of internal representations via downstream tasks \cite{baan2019realization, ettinger2018assessing, conneau2019unsupervised, nandakumar2019well, mccoy2019right}. One line of work analyzes word interactions in neural networks' internal gates as the composition signal \cite{saphra2020lstms, murdoch2018beyond}, extending the Contextual Decomposition algorithm proposed by \citet{jumelet2019analysing}. Another notable branch of work constructs synthetic datasets of small size to investigate compositionality in neural networks \cite{livska2018memorize, hupkes2018learning, baan2019realization}. Some work controls for word content, as we do, to study composition at the sentence level \cite{ettinger2018assessing,dasgupta2018evaluating}. We complement this work with a targeted and systematic study of phrase-level representations in transformers, with a focus on teasing apart lexical properties versus reflections of accurate compositional phrase meaning. 

Our work relates closely to classic work on two-word phrases, which have used methods like landmark tests \cite{kintsch2001predication,mitchell2008vector, mitchell2010composition}, or compared against distribution-based phrase representations \cite{baroni2010nouns,fyshe2015compositional}.
Our work also draws on work using correlation with similarity judgments \cite{finkelstein2001placing, gerz2016simverb, hill2015simlex, conneau2018senteval} and paraphrase classification \cite{ganitkevitch2013ppdb, wang2018glue, paws2019naacl, pawsx2019emnlp} to assess quality of models and representations. We build on this work by combining these methods together, applying them to a systematic analysis of transformers and their components, and introducing controlled variants of existing tasks to isolate accurate composition of phrase meaning from capturing of lexical information.

\section{Testing phrase meaning similarity} \label{sec:tasks}

Our methods begin with familiar approaches for assessing representations via meaning similarity: correlation with human phrase similarity judgments, and ability to identify paraphrases. The goal is to gauge the extent to which models arrive at representations reflecting the nuances of composed phrase meaning understood by humans. 
We draw on existing datasets, and begin by testing models on the original versions of these datasets---then we tease apart effects of word content from effects of more sophisticated meaning composition by introducing controlled variants of the datasets. The reasoning is that strong correlations with human similarity judgments, or strong paraphrase classification performance, could be influenced by artifacts that are not reflective of accurate phrase meaning composition per se. In particular, we may see strong performance simply on the basis of the amount of overlap in word content between phrases. To address this possibility, we create controlled datasets in which word overlap is no longer a cue to similarity. 

As a starting point we focus on two-word phrases, as these are the smallest phrasal unit and the most conducive to these types of lexical controls, and because this allows us to leverage larger amounts of annotated phrase similarity data.

\subsection{Phrase similarity correlation}

\begin{table}[t]
    \centering
    \begin{tabular}{cc}
    \thick
    \multicolumn{2}{c}{\textbf{Normal Examples}} \\
Source Phrase         & Target Phrase \& Score         \\ \thick
                      & ordinary citizen (0.724)     \\ \cline{2-2}
average person        & person average (0.518)       \\ \cline{2-2}
                      & country (0.255)           \\ \thick
    \multicolumn{2}{c}{\textbf{AB-BA Examples}} \\
Source Phrase         & Target Phrase \& Score         \\ \thick
law school            & school law (0.382)   \\ \cline{1-2}
adult female          & female adult (0.812)          \\ \cline{1-2}
arms control          & control arms (0.473)          \\ \thick
    \end{tabular}
\caption{Examples of correlation items. Numbers in parentheses are similarity scores between target phrase and source phrase. Upper half shows normal examples, and lower half shows controlled items.}
\label{tab:cor_example}
\end{table}

We first evaluate phrase representations by assessing their alignment with human judgments of phrase meaning similarity. For testing this correspondence, we use the \textbf{BiRD}~\cite{asaadi2019big} dataset. BiRD is a bigram relatedness dataset designed to evaluate composition, consisting of 3,345 bigram pairs (examples in Table~\ref{tab:cor_example}), with source phrases paired with numerous target phrases, and human-rated similarity scores ranging from 0 to 1.

In addition to testing on the full dataset, we design a controlled experiment to remove effects of word overlap, by filtering the dataset to pairs in which the two phrases consist of the same words. We refer to these pairs as ``AB-BA'' pairs (following terminology of the authors of the BiRD dataset), and show examples in the lower half of Table \ref{tab:cor_example}.

We run similarity tests as follows: given a model \emph{M} with layers \emph{L}, for $i$th layer $l_i \in L$ and a source-target phrase pair, we compute representations of source phrase $p^i_{rep}(\textrm{src})$ and target phrase $p^i_{rep}(\textrm{trg})$, where $rep$ is a representation type from Section \ref{representation}, and we compute their cosine cos($p^i_{rep}(\textrm{src}), p^i_{rep}(\textrm{trg})$). Pearson correlation $\bm{r_i}$ of layer $l_i$ is then computed between cosine and human-rated score for all source-target pairs. 

\subsection{Paraphrase classification}

\begin{table*}[ht]
    \centering
    \begin{tabular}{cc}
    \thick
    \multicolumn{2}{c}{\textbf{Normal Examples}} \\
Source Phrase         & Target Phrase        \\ \thick
 \multirow{4}{*}{are crucial}                       & is absolutely vital (pos) \\ \cline{2-2}
       &  was a matter of concern (neg)       \\ \cline{2-2}
       &  is an essential part (pos)       \\ \cline{2-2}
       &  are exacerbating (neg)       \\ \thick
    \multicolumn{2}{c}{\textbf{Controlled Examples}} \\
Source Phrase         & Target Phrase         \\ \thick
  \multirow{2}{*}{communication infrastructure}           & telecommunications infrastructure (pos)  \\ \cline{2-2}
          & data infrastructure (neg)          \\ \thick
    \end{tabular}
\caption{Examples of classification items. Classification labels between target phrase and source phrase are in parentheses. Upper half shows normal examples, and lower half shows controlled items.}
\label{tab:clf_example}
\end{table*}

We further investigate the nature of phrase representations by testing their capacity to support binary paraphrase classification. This test allows us to explore whether we will see better alignment with human judgments of meaning similarity if we use more complicated operations than cosine similarity comparison. For the classification tasks, we draw on \textbf{PPDB 2.0} \cite{pavlick2015ppdb}, a widely-used database consisting of paraphrases with scores generated by a regression model. To formulate our binary classification task, after filtering out low-quality paraphrases (discussed in Section \ref{setup}), we use phrase pairs (source phrase, target phrase) from PPDB as positive pairs, and randomly sample phrases from the complete PPDB dataset to form negative pairs (source phrase, random phrase).  

Because word overlap is also a likely cue for paraphrase classification, we filter to a controlled version of this dataset as well, as illustrated in Table~\ref{tab:clf_example}. We formulate the controlled experiment here as holding word overlap between source phrase and target phrase to be exactly 50\% for both positive and negative samples. Our choice of 50\% word overlap in this case is necessary for construction of a sufficiently large, balanced classification dataset (AB-BA pairs in PPDB are too few to support classifier training, and AB-BA pairs are more likely to be non-paraphrases). Note, however, that by controlling word overlap to be exactly 50\% for all phrase pairs, we still hold constant the \emph{amount} of word overlap between phrases, which is the cue that we wish to remove. As an additional control, each source phrase is paired with an equal number of paraphrases and non-paraphrases, to avoid the classifier inferring labels based on phrase identity.

Formally, for each model layer $l_i$ and representation type $rep$, we train
\begin{equation*}
    \textrm{CLF}^i_{rep} = \textrm{MLP}([\bm{pair^i_{rep}}])
\end{equation*}
where $\bm{pair^i_{rep}}$ represents embedding concatenations of each source phrase and target phrase:
\begin{equation*}
    \bm{pair^i_{rep}} = [\bm{p^i_{rep}(src)}; \bm{p^i_{rep}(trg)}]
\end{equation*}
The classifier is trained on binary classification of whether concatenated inputs represent paraphrases. 


\section{Representation types}
\label{representation}

A variety of approaches have been taken for representing sentences and phrases when all tokens output contextualized representations, as in our tested transformers. To clarify the phrasal information present in different forms of phrase representation, we experiment with a number of different combinations of token embeddings as representation types. 

Formally, let [$T_0, \cdots, T_k$] be an input sequence of length $k+1$, with corresponding embeddings at $i$th layer [$\bm{e_0^i}, \cdots, \bm{e_k^i}$]. Assume the phrase spans the sequence [$a, b$], where $0 \leq a \leq b \leq k$. Because two-word phrases are atypical inputs for these models, we experiment both with inputs of the two-word phrases alone (``phrase-only''), as well as inputs with the phrases embedded in sentences (``context-available''). This is illustrated in Figure~\ref{fig:tok_pos} along with phrase representation types.


We test the following forms of phrase representation, drawn from each model and layer separately: 

\begin{figure}[t]
    \centering
    \includegraphics[width=0.47\textwidth]{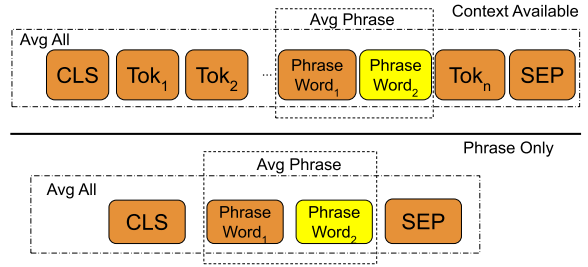}
    \caption{Example input sequences (BERT format). CLS is a special token at beginning of sequence. Tokens in yellow correspond to Head-Word. Avg-Phrase contains element-wise average of phrase word embeddings. Avg-All averages embeddings of all tokens.}
    \label{fig:tok_pos}
\end{figure}

\begin{figure*}[ht]
    \centering
    \includegraphics[width=1\textwidth]{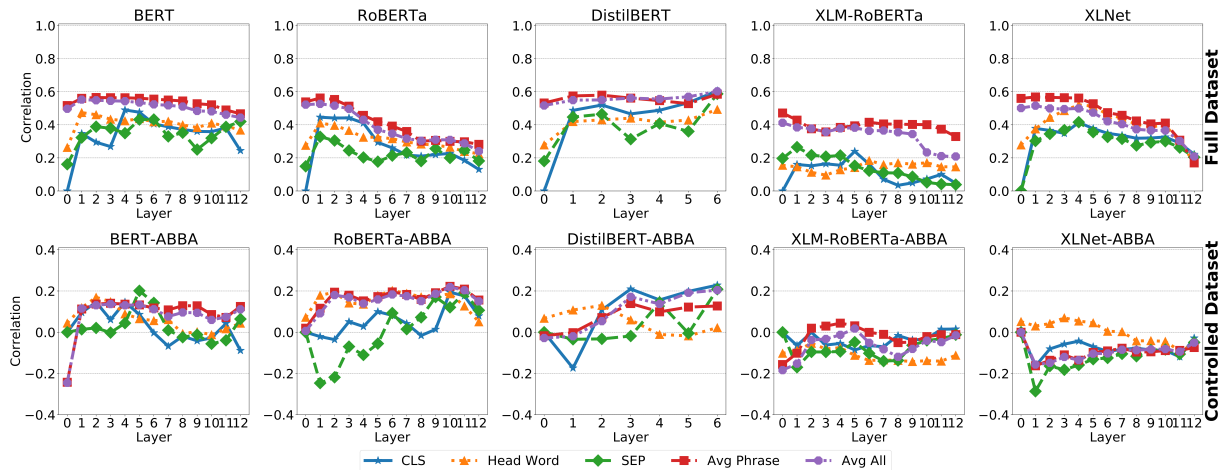}
    \caption{Correlation on BiRD dataset, phrase-only input setting. First row shows results on full dataset, and second row on controlled AB-BA pairs. Layer 0 corresponds to input embeddings passing to the model.}
    \label{fig:correlation_exp}
\end{figure*}

\paragraph{CLS} Depending on specific models, this special token can be the first or last token of the input sequence (i.e. $\bm{e_0^i}$ or $\bm{e_k^i}$). In many applications, this token is used to represent the full input sequence.

\paragraph{Head-Word} In each phrase, the head word is the semantic center the phrase. For instance, in the phrase ``public service'', ``service'' is the head word, expressing the central meaning of the phrase, while ``public'' is a modifier. Because phrase heads are not annotated in our datasets, we approximate the head by taking the embedding of the final word of the phrase. This representation is proposed as a potential representation of the whole phrase, if information is being composed into a central word:
\begin{equation*}
    \bm{p_{hw}^i} = \bm{e_b^i}
\end{equation*}
\paragraph{Avg-Phrase} For this representation type we average the embeddings of the tokens in the target phrase (dashed box in Figure \ref{fig:tok_pos}). This type of averaging of token embeddings is a common means of aggregate representation \cite{wieting2015towards}.
\begin{equation*}
    \bm{p_{ap}^i} = \frac{1}{b - a + 1} \sum_{x=a}^b \bm{e_x^i}
\end{equation*}
\paragraph{Avg-All} Expanding beyond the tokens in ``Avg-Phrase'', this representation averages embeddings from the full input sequence.
\begin{equation*}
    \bm{p_{aa}^i} = \frac{1}{k + 1} \sum_{x=0}^k \bm{e_x^i}
\end{equation*}
\paragraph{SEP} With some variation between models, the SEP token is typically a separator for distinguishing input sentences, and is often the last token ($\bm{e_k^i}$) or second to last token ($\bm{e_{k-1}^i}$) of a sequence.

\section{Experimental setup}
\label{setup}

 Embeddings of each token are obtained by feeding input sequences through pre-trained contextual encoders. We investigate the ``base'' version of five transformers: BERT \cite{devlin2019bert}, RoBERTa \cite{liu2019roberta}, DistilBERT \cite{sanh2019distilbert}, XLNet \cite{yang2019xlnet} and XLM-RoBERTa \cite{conneau2019unsupervised}. For the models analyzed in this paper, we are using the implementation of \citet{Wolf2019HuggingFacesTS},\footnote{https://github.com/huggingface/transformers} which is based on PyTorch \cite{paszke2019pytorch}.  
 
 For correlation analysis, we first use the complete BiRD dataset, consisting of 3,345 phrase pairs.\footnote{http://saifmohammad.com/WebPages/BiRD.html} We then test with our controlled subset of the data, consisting of 410 AB-BA pairs. For classification tasks, we first do preprocessing on PPDB 2.0,\footnote{http://paraphrase.org} filtering out pairs containing hyperlinks, non-alphabetical symbols, and trivial paraphrases based on abbreviation or tense change. For our initial classification test, we use 13,050 source-target phrase pairs (of varying word overlap) from this preprocessed dataset. We then test with our controlled dataset, consisting of 11,770 source-target phrase pairs (each with precisely 50\% word overlap). For each paraphrase classification task, 25\% of selected data is reserved for testing. We use a multi-layer perceptron classifier with a single hidden layer of size 256 with ReLU activation, and a softmax layer to generate binary labels. We use a relatively simple classifier following the reasoning of \citet{adi2016fine}, that this allows examination of how easily extractable information is in these representations. 
 
For both correlation and classification tasks, we experiment with phrase-only inputs and context-available (full-sentence) inputs. To obtain sentence contexts, we search for instances of source phrases in a Wikipedia dump, and extract sentences containing them. For a given phrase pair, target phrases are embedded in the same sentence context as the source phrase, to avoid effects of varying sentence position between phrases of a given pair. \footnote{Because context sentences are extracted based on source phrases, our use of the same context for source and target phrases can give rise to unnatural contextual fit for target phrases. We consider this acceptable for the sake of controlling sentence position---and if anything, differences in contextual fit may aid models in distinguishing more and less similar phrases. The slight boost observed on the full datasets (for Avg-Phrase) suggests that the sentence contexts do provide the intended benefit from using input of a more natural size.}

\section{Results}

\begin{figure*}[ht]
    \centering
    \includegraphics[width=1\textwidth]{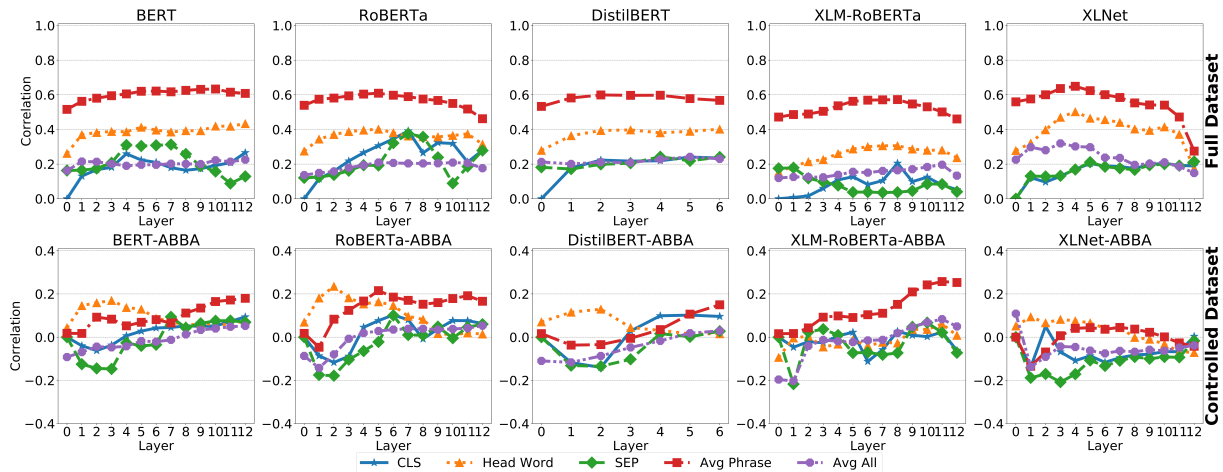}
    \caption{Correlation on BiRD dataset with phrases embedded in sentence context (context-available input setting).}
    \label{fig:correlation_in_sents_exp}
\end{figure*}

\subsection{Similarity correlation}
\paragraph{Full dataset}  \ The top row of Figure \ref{fig:correlation_exp} shows correlation results on the full BiRD dataset for all models, layers, and representation types,  with phrase-only inputs. Among representation types, Avg-Phrase and Avg-All consistently achieve the highest correlations across models and layers. In all models but DistilBERT, correlation of Avg-Phrase and Avg-All peaks at layer 1 and decreases in subsequent layers with minor fluctuations. Head-Word and SEP both show weaker, but non-trivial, correlations. The CLS token is of note with a consistent rapid rise as layers progress, suggesting that it quickly takes on properties of the words of the phrase. For all models but DistilBERT, CLS token correlations peak in middle layers and then decline. 

Model-wise, XLM-RoBERTa shows the weakest overall correlations, potentially due to the fact that it is trained to infer input language and to handle multiple languages. BERT retains fairly consistent correlations across layers, while RoBERTa and XLNet show rapid declines as layers progress, suggesting that these models increasingly incorporate information that deviates from human intuitions about phrase smilarity. DistilBERT, despite being of smaller size, demonstrates competitive correlation. The CLS token in DistilBERT is notable for its continuing rise in correlation strength across layers. This suggests that DistilBERT in particular makes use of the CLS token to encode phrase information, and unlike other models, its representations retain the relevant properties to the final layer.

\paragraph{Controlled dataset} 
Turning to our controlled AB-BA dataset, we examine the extent to which the above correlations indicate sophisticated phrasal composition versus effective encoding of information about phrases' component words. The bottom row of Figure \ref{fig:correlation_exp} shows the correlations on this controlled subset. We see that performance of all models drops significantly, often with roughly zero correlation. Avg-All and Avg-Phrase no longer dominate the correlations, suggesting that these representations capture word information, but not higher-level compositional information. XLM-RoBERTa and XLNet show particularly low correlations, suggesting heavier reliance on word content.  Notably, the CLS tokens in RoBERTa  and DistilBERT stand out with comparatively strong correlations in later layers. This suggests that the rise that we see in CLS correlations for DistilBERT in particular may correspond to some real compositional signal in this token, and for this model the CLS token may in fact correspond to something more like a representation of the meaning of the full input sequence. The Avg-Phrase representation for RoBERTa also makes a comparatively strong showing.

\begin{figure*}[ht]
    \centering
    \includegraphics[width=1\textwidth]{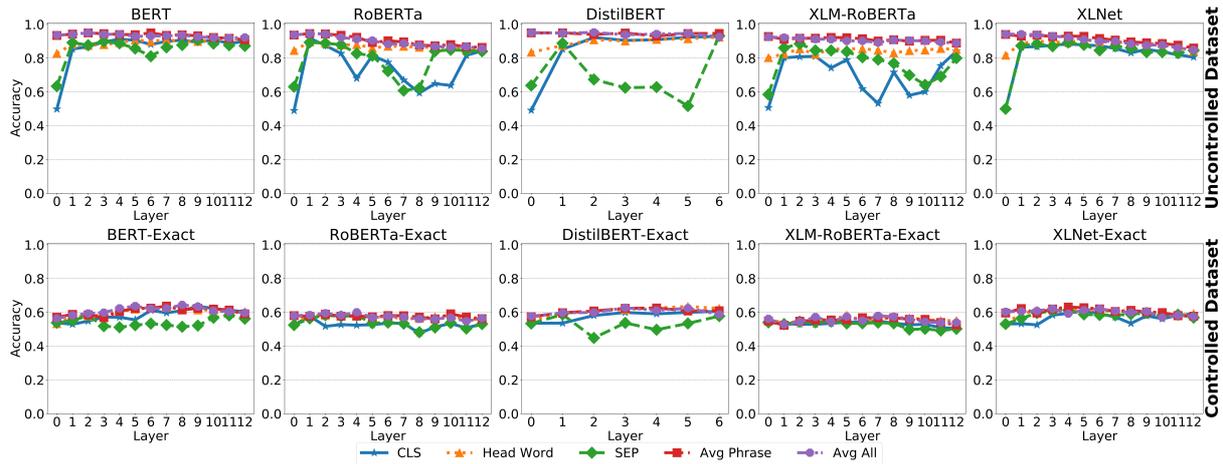}
    \caption{Classification accuracy on PPDB dataset (phrase-only input setting). First row shows classification accuracy on original dataset, and second row shows accuracy on controlled dataset.}
    \label{fig:classification_exp}
\end{figure*}

\begin{figure*}[ht]
    \centering
    \includegraphics[width=1\textwidth]{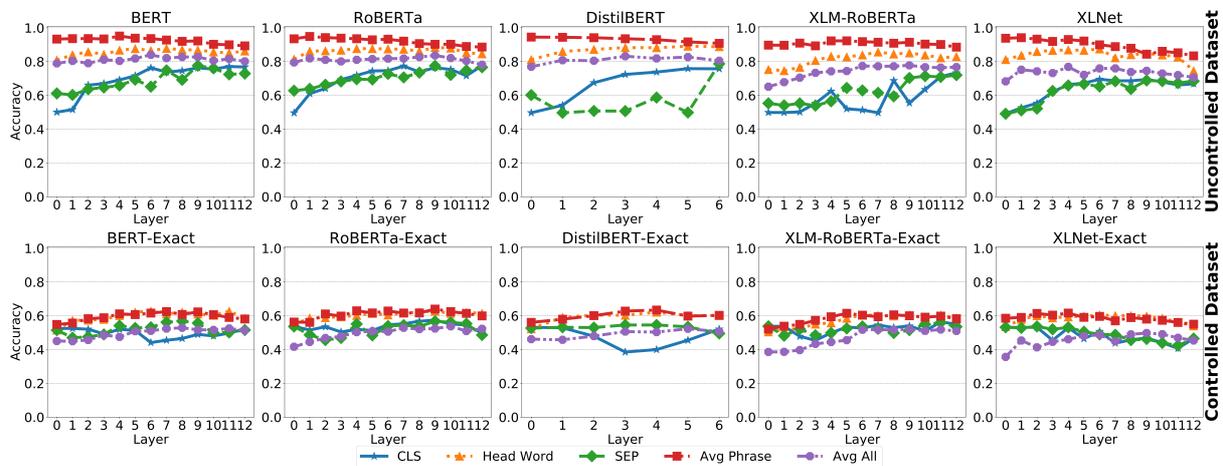}
    \caption{Classification accuracy on PPDB dataset with phrases embedded in sentence context. First row shows classification accuracy on original dataset, and second row shows accuracy on controlled dataset.}
    \label{fig:ppdb_in_sent}
\end{figure*}


\paragraph{Including sentence context} Figure \ref{fig:correlation_in_sents_exp} shows the correlations when target phrases are embedded as part of a sentence context, rather than in isolation. As can be expected, Avg-Phrase is now consistently the highest in correlation on the full dataset---other tokens are presumably more impacted by the presence of additional words in the context. We also see that the Avg-Phrase correlations no longer drop so dramatically in later layers, suggesting that when given full sentence inputs, models retain more word properties in later layers than when given only phrases.  This general trend holds also for Avg-All and Head-Word representations.

In the AB-BA setting, we see that presence of context does boost overall correlation with human judgment. Of note is XLM-RoBERTa's Avg-Phrase, which without sentence context has zero correlation in the AB-BA setting, but which with sentence context reaches our highest observed AB-BA correlations in its final layers. However, even with context, the strongest correlation across models is still less than 0.3. It is still the case, then, that correlation on the controlled data degrades significantly relative to the full dataset. This indicates that even when phrases are input within sentence contexts, phrase representations in transformers reflect heavy reliance on word content, largely missing additional nuances of compositional phrase meaning. 

\subsection{Paraphrase classification}
\paragraph{Full dataset} Results for our full paraphrase classification dataset, with phrase-only inputs, are shown in the top row of Figure \ref{fig:classification_exp}. Accuracies are overall very high, and we see generally similar patterns to the correlation tasks. Best accuracy is achieved by using Avg-Phrase and Avg-All representations. RoBERTa, XLM-RoBERTa, and XLNet show decreasing correlations for top-performing representations in later layers, while BERT and DistilBERT remain more consistent across layers. Performance of CLS requires a few layers to peak, with top performance around middle layers, and in some models shows poor performance in later layers. SEP shows unstable performance compared to other representations, especially in DistilBERT and RoBERTa.
 
\paragraph{Controlled dataset} The bottom row of Figure \ref{fig:classification_exp} shows classification accuracy when word overlap is held constant. Consistent with the drop in correlations on the controlled AB-BA experiments above, classification performance of all models drops down to only slightly above chance performance of 50\%. This suggests that the high classification performance on the full dataset relies largely on word overlap information, and that there is little higher-level phrase meaning information to aid classification in the absence of the overlap cue. We see in some cases a very slight trend such that classification accuracy increases a bit toward middle layers---so to the extent that there is any compositional phrase information being captured, it may increase within representations in the middle layers. Overall, the consistency of these results with those of the correlation analysis suggests that the apparent lack of accurate compositional meaning information in our tested phrase representations is not simply a result of cosine correlations being inappropriate for picking up on correspondences.

\paragraph{Including sentence context} Figure \ref{fig:ppdb_in_sent} shows the classification results for representations of phrases embedded in sentence contexts. The patterns largely align with our observations from the correlation task. Performance on the full dataset is still high, with Avg-Phrase now showing consistently highest performance, being least influenced by the presence of new context words. In the controlled setting, we see the same substantial drop in performance relative to the full dataset---there is very slight improvement over the phrase-only representations, but the highest accuracy among all models is still around 0.6. Thus, the inclusion of sentence context again does not provide any additional evidence for sophisticated compositional meaning information in the tested phrase representations.


\section{Qualitative analysis: sense disambiguation} \label{landmark}

The above analyses rely on testing models' sensitivity to meaning similarity between two phrases. In this section we complement these analyses with another test aimed at assessing phrasal composition: testing models' ability to select the correct senses of polysemous words in a composed phrase, as proposed by~\citet{kintsch2001predication}. Each test item consists of a) a central verb, b) two subject-verb phrases that pick out different senses of the verb, and c) two \emph{landmark} words, each associating with one of the target senses of the verb. Table \ref{tab:landmark} shows an example with central verb ``ran'' and phrases ``horse ran''/ ``color ran''. The corresponding landmark words are ``gallop'', which associates with ``horse ran'', and ``dissolve'', which associates with ``color ran''. The reasoning is that composition should select the correct verb meaning, shifting representations of the central verbs---and of the phrase as a whole---toward landmarks with closer meaning. For this example, models should produce phrase embeddings such that ``horse ran'' is closer to ``gallop'' and ``color ran'' is closer to ``dissolve''.  We use the items introduced in \citet{kintsch2001predication}, which consist of a total of 4 sets of landmark tests. We feed landmarks and phrases respectively through each transformer, without context, to generate corresponding representations $p^i_{rep}$ for each layer $l_i$ and representation type $rep$. Cosine similarity between each phrase-landmark pair is computed and compared against expected similarities.




\begin{table}[]
\centering
\begin{tabular}{c|c|c}
\thick
  & \emph{horse ran} & \emph{color ran} \\ \thick
\emph{gallop}         & POS     &  NEG         \\ \hline
\emph{dissolve}       &    NEG       & POS     \\ \thick
\end{tabular}
\caption{An example of landmark experiment of verb "run". Representations are expected to have higher cosine similarities between phrase and landmark word that are marked ``POS''.}
\label{tab:landmark}
\end{table}

\begin{figure*}[ht]
    \centering
    \includegraphics[width=1\textwidth]{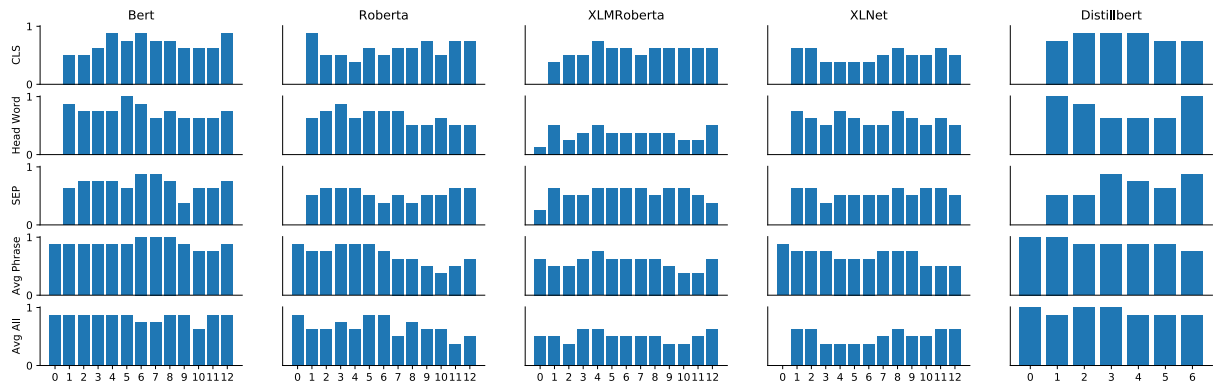}
    \caption{Landmark experiments. Y-axis denotes the percentage of samples that are shifted towards the correct landmark words in each layer. Missing bars occur when representations are independent of input at layer 0, such that cosine similarity between phrases and landmarks will always be 1.}
    \label{fig:landmark_exp}
\end{figure*}

Figure \ref{fig:landmark_exp} shows the percentage of phrases that fall closer to the correct landmark word than to the incorrect one, averaged over 16 phrase-landmark word pairs. We see strong overall performance across models, suggesting that the information needed for this task is successfully captured by these models' representations. Additionally, we see that the patterns largely mirror the results above for correlation and classification on uncontrolled datasets. Particularly, Avg-Phrase and Avg-All show comparatively strong performance across models. RoBERTa and XLNet show stronger performance in early layers, dropping off in later layers, while BERT and DistilBERT show more consistency across layers. XLM-RoBERTa and XLNet show lower performance overall. 

For this verb sense disambiguation analysis, the Head-Word token is of note because it corresponds to the central verb of interest, so its sense can only be distinguished by its combination with the other word of the phrase. XLM-RoBERTa has the weakest performance with Head-Word, while BERT and DistilBERT demonstrate strong disambiguation with this token. As for the CLS token, RoBERTa  produces the highest quality representation at layer 1, and BERT outperforms other models starting from layer 6, with DistilBERT also showing strong performance across layers.

Notably, the observed parallels to our correlation and classification results are in alignment with the uncontrolled rather than the controlled versions of those tests. So while these parallels lend further credence to the general observations that we make about phrase representation patterns across models, layers, and representation types, it is worth noting that these landmark composition tests may be susceptible to lexical effects similar to those controlled for above. Since these test items are too few to filter with the above methods, we leave in-depth investigation of this question to future work. 

\section{Discussion}
The analyses reported above yield two primary takeaways. First, they shed light on the nature of these models' phrase representations, and the extent to which they reflect word content versus phrasal composition. At many points in these models there is non-trivial alignment with human judgments of phrase similarity, paraphrase classification, and verb sense selection. However, when we control our correlation and classification tests to remove the cue of word overlap, we see little evidence that the representations reflect sophisticated phrase composition beyond what can be gleaned from word content. While we see strong performance on classic sense selection items designed to test phrase composition, the observed results largely parallel those from the uncontrolled versions of the correlation and classification analyses, suggesting that success on this landmark test may reflect lexical properties more than sophisticated composition. Given the importance of systematic meaning composition for robust and flexible language understanding, based on these results we predict that we will see corresponding weaknesses as more tests emerge for these models' handling of subtle meaning differences in downstream tasks. 


Our systematic examination of models, layers and representation types yields a second takeaway in the form of practical implications for selecting and extracting representations from these models. For faithful representations of word content, Avg-Phrase is generally the strongest candidate. If only the phrase is embedded, drawing from earlier layers is best in RoBERTa, XLM-RoBERTa, and XLNet, while middle layers are better in BERT, and later layers in DistilBERT. If the phrase is input as part of a sentence, middle layers are generally best across models. Though the CLS token is often interpreted to represent a full input sequence, we find it to be a poor phrase representation even with phrase-only input, with the notable exception of the final layer of DistilBERT.

As for representations that reflect true phrase meaning composition, we have established that such representations may not currently be available in these models. However, to the extent that we do see weak evidence of potential compositional meaning sensitivity, this appears to be strongest in DistilBERT's CLS token in final layers, in RoBERTa's Avg-Phrase representation in later layers, and in XLM-RoBERTa's Avg-Phrase representation from later layers \emph{only} when the phrase is contained within a sentence context.  


\section{Conclusions and future directions}

We have systematically investigated the nature of phrase representations in state-of-the-art transformers.
Teasing apart sensitivity to word content versus phrase meaning composition, we find strong sensitivity across models when it comes to word content encoding, but little evidence of sophisticated phrase composition. The observed sensitivity patterns across models, layers, and representation types shed light on practical considerations for extracting phrase representations from these models. 

Future work can apply these tests to a broader range of models, and continue to develop controlled tests that target encoding of complex compositional meanings, both for two-word phrases and for larger meaning units. We hope that our findings will stimulate further work on leveraging the power of these generalized transformers while improving their capacity to capture compositional meaning.


\section*{Acknowledgments}
We would like to thank three anonymous reviewers for valuable questions and suggestions for improving this paper. We also thank members of the University of Chicago CompLing Lab, and the Toyota Technological Institute at Chicago, for helpful comments and feedback on this work. This material is based upon work supported by the National Science Foundation under Award No. 1941160.

\bibliography{emnlp2020}

\begin{thebibliography}{61}
\expandafter\ifx\csname natexlab\endcsname\relax\def\natexlab#1{#1}\fi

\bibitem[{Adi et~al.(2016)Adi, Kermany, Belinkov, Lavi, and
  Goldberg}]{adi2016fine}
Yossi Adi, Einat Kermany, Yonatan Belinkov, Ofer Lavi, and Yoav Goldberg. 2016.
\newblock Fine-grained analysis of sentence embeddings using auxiliary
  prediction tasks.
\newblock \emph{arXiv preprint arXiv:1608.04207}.

\bibitem[{Asaadi et~al.(2019)Asaadi, Mohammad, and Kiritchenko}]{asaadi2019big}
Shima Asaadi, Saif Mohammad, and Svetlana Kiritchenko. 2019.
\newblock Big bird: A large, fine-grained, bigram relatedness dataset for
  examining semantic composition.
\newblock In \emph{Proceedings of the 2019 Conference of the North American
  Chapter of the Association for Computational Linguistics: Human Language
  Technologies, Volume 1 (Long and Short Papers)}, pages 505--516.

\bibitem[{Baan et~al.(2019)Baan, Leible, Nikolaus, Rau, Ulmer, Baumg{\"a}rtner,
  Hupkes, and Bruni}]{baan2019realization}
Joris Baan, Jana Leible, Mitja Nikolaus, David Rau, Dennis Ulmer, Tim
  Baumg{\"a}rtner, Dieuwke Hupkes, and Elia Bruni. 2019.
\newblock On the realization of compositionality in neural networks.
\newblock In \emph{Proceedings of the 2019 ACL Workshop BlackboxNLP: Analyzing
  and Interpreting Neural Networks for NLP}, pages 127--137.

\bibitem[{Bacon and Regier(2019)}]{bacon2019does}
Geoff Bacon and Terry Regier. 2019.
\newblock Does bert agree? evaluating knowledge of structure dependence through
  agreement relations.
\newblock \emph{arXiv preprint arXiv:1908.09892}.

\bibitem[{Baroni and Zamparelli(2010)}]{baroni2010nouns}
Marco Baroni and Roberto Zamparelli. 2010.
\newblock Nouns are vectors, adjectives are matrices: Representing
  adjective-noun constructions in semantic space.
\newblock In \emph{Proceedings of the 2010 Conference on Empirical Methods in
  Natural Language Processing}, pages 1183--1193. Association for Computational
  Linguistics.

\bibitem[{Chowdhury and Zamparelli(2018)}]{chowdhury2018rnn}
Shammur~Absar Chowdhury and Roberto Zamparelli. 2018.
\newblock {RNN} simulations of grammaticality judgments on long-distance
  dependencies.
\newblock In \emph{Proceedings of the 27th international conference on
  computational linguistics}, pages 133--144.

\bibitem[{Clark et~al.(2019)Clark, Khandelwal, Levy, and
  Manning}]{clark2019does}
Kevin Clark, Urvashi Khandelwal, Omer Levy, and Christopher~D Manning. 2019.
\newblock What does bert look at? an analysis of bert’s attention.
\newblock In \emph{Proceedings of the 2019 ACL Workshop BlackboxNLP: Analyzing
  and Interpreting Neural Networks for NLP}, pages 276--286.

\bibitem[{Conneau et~al.(2019)Conneau, Khandelwal, Goyal, Chaudhary, Wenzek,
  Guzm{\'a}n, Grave, Ott, Zettlemoyer, and Stoyanov}]{conneau2019unsupervised}
Alexis Conneau, Kartikay Khandelwal, Naman Goyal, Vishrav Chaudhary, Guillaume
  Wenzek, Francisco Guzm{\'a}n, Edouard Grave, Myle Ott, Luke Zettlemoyer, and
  Veselin Stoyanov. 2019.
\newblock Unsupervised cross-lingual representation learning at scale.
\newblock \emph{arXiv preprint arXiv:1911.02116}.

\bibitem[{Conneau and Kiela(2018)}]{conneau2018senteval}
Alexis Conneau and Douwe Kiela. 2018.
\newblock Senteval: An evaluation toolkit for universal sentence
  representations.
\newblock In \emph{Proceedings of the Eleventh International Conference on
  Language Resources and Evaluation (LREC 2018)}.

\bibitem[{Conneau et~al.(2018)Conneau, Kruszewski, Lample, Barrault, and
  Baroni}]{conneau2018you}
Alexis Conneau, Germ{\'a}n Kruszewski, Guillaume Lample, Lo{\"\i}c Barrault,
  and Marco Baroni. 2018.
\newblock What you can cram into a single \$ \&!\#* vector: Probing sentence
  embeddings for linguistic properties.
\newblock In \emph{Proceedings of the 56th Annual Meeting of the Association
  for Computational Linguistics (Volume 1: Long Papers)}, pages 2126--2136.

\bibitem[{Dasgupta et~al.(2018)Dasgupta, Guo, Stuhlm{\"u}ller, Gershman, and
  Goodman}]{dasgupta2018evaluating}
Ishita Dasgupta, Demi Guo, Andreas Stuhlm{\"u}ller, Samuel~J Gershman, and
  Noah~D Goodman. 2018.
\newblock Evaluating compositionality in sentence embeddings.
\newblock \emph{arXiv preprint arXiv:1802.04302}.

\bibitem[{Devlin et~al.(2019)Devlin, Chang, Lee, and
  Toutanova}]{devlin2019bert}
Jacob Devlin, Ming-Wei Chang, Kenton Lee, and Kristina Toutanova. 2019.
\newblock Bert: Pre-training of deep bidirectional transformers for language
  understanding.
\newblock In \emph{Proceedings of the 2019 Conference of the North American
  Chapter of the Association for Computational Linguistics: Human Language
  Technologies, Volume 1 (Long and Short Papers)}, pages 4171--4186.

\bibitem[{Dyer et~al.(2016)Dyer, Kuncoro, Ballesteros, and
  Smith}]{dyer2016recurrent}
Chris Dyer, Adhiguna Kuncoro, Miguel Ballesteros, and Noah~A Smith. 2016.
\newblock Recurrent neural network grammars.
\newblock In \emph{Proceedings of the 2016 Conference of the North American
  Chapter of the Association for Computational Linguistics: Human Language
  Technologies}, pages 199--209.

\bibitem[{Ettinger et~al.(2018)Ettinger, Elgohary, Phillips, and
  Resnik}]{ettinger2018assessing}
Allyson Ettinger, Ahmed Elgohary, Colin Phillips, and Philip Resnik. 2018.
\newblock Assessing composition in sentence vector representations.
\newblock In \emph{Proceedings of the 27th International Conference on
  Computational Linguistics}, pages 1790--1801.

\bibitem[{Ettinger et~al.(2016)Ettinger, Elgohary, and
  Resnik}]{ettinger2016probing}
Allyson Ettinger, Ahmed Elgohary, and Philip Resnik. 2016.
\newblock Probing for semantic evidence of composition by means of simple
  classification tasks.
\newblock In \emph{Proceedings of the 1st Workshop on Evaluating Vector-Space
  Representations for NLP}, pages 134--139.

\bibitem[{Finkelstein et~al.(2001)Finkelstein, Gabrilovich, Matias, Rivlin,
  Solan, Wolfman, and Ruppin}]{finkelstein2001placing}
Lev Finkelstein, Evgeniy Gabrilovich, Yossi Matias, Ehud Rivlin, Zach Solan,
  Gadi Wolfman, and Eytan Ruppin. 2001.
\newblock Placing search in context: The concept revisited.
\newblock In \emph{Proceedings of the 10th international conference on World
  Wide Web}, pages 406--414.

\bibitem[{Futrell et~al.(2019)Futrell, Wilcox, Morita, Qian, Ballesteros, and
  Levy}]{futrell2019neural}
Richard Futrell, Ethan Wilcox, Takashi Morita, Peng Qian, Miguel Ballesteros,
  and Roger Levy. 2019.
\newblock Neural language models as psycholinguistic subjects: Representations
  of syntactic state.
\newblock In \emph{Proceedings of the 2019 Conference of the North American
  Chapter of the Association for Computational Linguistics: Human Language
  Technologies, Volume 1 (Long and Short Papers)}, pages 32--42.

\bibitem[{Fyshe et~al.(2015)Fyshe, Wehbe, Talukdar, Murphy, and
  Mitchell}]{fyshe2015compositional}
Alona Fyshe, Leila Wehbe, Partha Talukdar, Brian Murphy, and Tom Mitchell.
  2015.
\newblock A compositional and interpretable semantic space.
\newblock In \emph{Proceedings of the 2015 conference of the north american
  chapter of the association for computational linguistics: Human language
  technologies}, pages 32--41.

\bibitem[{Ganitkevitch et~al.(2013)Ganitkevitch, Van~Durme, and
  Callison-Burch}]{ganitkevitch2013ppdb}
Juri Ganitkevitch, Benjamin Van~Durme, and Chris Callison-Burch. 2013.
\newblock Ppdb: The paraphrase database.
\newblock In \emph{Proceedings of the 2013 Conference of the North American
  Chapter of the Association for Computational Linguistics: Human Language
  Technologies}, pages 758--764.

\bibitem[{Gerz et~al.(2016)Gerz, Vuli{\'c}, Hill, Reichart, and
  Korhonen}]{gerz2016simverb}
Daniela Gerz, Ivan Vuli{\'c}, Felix Hill, Roi Reichart, and Anna Korhonen.
  2016.
\newblock Simverb-3500: A large-scale evaluation set of verb similarity.
\newblock In \emph{Proceedings of the 2016 Conference on Empirical Methods in
  Natural Language Processing}, pages 2173--2182.

\bibitem[{Gulordava et~al.(2018)Gulordava, Bojanowski, Grave, Linzen, and
  Baroni}]{gulordava2019colorless}
Kristina Gulordava, Piotr Bojanowski, Edouard Grave, Tal Linzen, and Marco
  Baroni. 2018.
\newblock Colorless green recurrent networks dream hierarchically.
\newblock In \emph{Proceedings of the 2018 Conference of the North American
  Chapter of the Association for Computational Linguistics: Human Language
  Technologies}.

\bibitem[{Hewitt and Manning(2019)}]{hewitt2019structural}
John Hewitt and Christopher~D Manning. 2019.
\newblock A structural probe for finding syntax in word representations.
\newblock In \emph{Proceedings of the 2019 Conference of the North American
  Chapter of the Association for Computational Linguistics: Human Language
  Technologies, Volume 1 (Long and Short Papers)}, pages 4129--4138.

\bibitem[{Hill et~al.(2015)Hill, Reichart, and Korhonen}]{hill2015simlex}
Felix Hill, Roi Reichart, and Anna Korhonen. 2015.
\newblock Simlex-999: Evaluating semantic models with (genuine) similarity
  estimation.
\newblock \emph{Computational Linguistics}, 41(4):665--695.

\bibitem[{Hupkes et~al.(2018)Hupkes, Singh, Korrel, Kruszewski, and
  Bruni}]{hupkes2018learning}
Dieuwke Hupkes, Anand Singh, Kris Korrel, German Kruszewski, and Elia Bruni.
  2018.
\newblock Learning compositionally through attentive guidance.
\newblock \emph{arXiv preprint arXiv:1805.09657}.

\bibitem[{Jawahar et~al.(2019)Jawahar, Sagot, and Seddah}]{jawahar2019does}
Ganesh Jawahar, Beno{\^\i}t Sagot, and Djam{\'e} Seddah. 2019.
\newblock What does bert learn about the structure of language?
\newblock In \emph{Proceedings of the 57th Annual Meeting of the Association
  for Computational Linguistics}, pages 3651--3657.

\bibitem[{Jumelet et~al.(2019)Jumelet, Zuidema, and
  Hupkes}]{jumelet2019analysing}
Jaap Jumelet, Willem Zuidema, and Dieuwke Hupkes. 2019.
\newblock Analysing neural language models: Contextual decomposition reveals
  default reasoning in number and gender assignment.
\newblock In \emph{Proceedings of the 23rd Conference on Computational Natural
  Language Learning (CoNLL)}, pages 1--11.

\bibitem[{Kim et~al.(2019)Kim, Rush, Yu, Kuncoro, Dyer, and
  Melis}]{kim2019unsupervised}
Yoon Kim, Alexander~M Rush, Lei Yu, Adhiguna Kuncoro, Chris Dyer, and G{\'a}bor
  Melis. 2019.
\newblock Unsupervised recurrent neural network grammars.
\newblock In \emph{Proceedings of the 2019 Conference of the North American
  Chapter of the Association for Computational Linguistics: Human Language
  Technologies, Volume 1 (Long and Short Papers)}, pages 1105--1117.

\bibitem[{Kintsch(2001)}]{kintsch2001predication}
Walter Kintsch. 2001.
\newblock Predication.
\newblock \emph{Cognitive science}, 25(2):173--202.

\bibitem[{Klafka and Ettinger(2020)}]{klafka2020spying}
Josef Klafka and Allyson Ettinger. 2020.
\newblock Spying on your neighbors: Fine-grained probing of contextual
  embeddings for information about surrounding words.
\newblock \emph{arXiv preprint arXiv:2005.01810}.

\bibitem[{Linzen et~al.(2016)Linzen, Dupoux, and
  Goldberg}]{linzen2016assessing}
Tal Linzen, Emmanuel Dupoux, and Yoav Goldberg. 2016.
\newblock Assessing the ability of {LSTM}s to learn syntax-sensitive
  dependencies.
\newblock \emph{Transactions of the Association for Computational Linguistics},
  4:521--535.

\bibitem[{Li{\v{s}}ka et~al.(2018)Li{\v{s}}ka, Kruszewski, and
  Baroni}]{livska2018memorize}
Adam Li{\v{s}}ka, Germ{\'a}n Kruszewski, and Marco Baroni. 2018.
\newblock Memorize or generalize? searching for a compositional rnn in a
  haystack.
\newblock \emph{arXiv preprint arXiv:1802.06467}.

\bibitem[{Liu et~al.(2019{\natexlab{a}})Liu, Gardner, Belinkov, Peters, and
  Smith}]{liu2019linguistic}
Nelson~F Liu, Matt Gardner, Yonatan Belinkov, Matthew~E Peters, and Noah~A
  Smith. 2019{\natexlab{a}}.
\newblock Linguistic knowledge and transferability of contextual
  representations.
\newblock In \emph{Proceedings of the 2019 Conference of the North American
  Chapter of the Association for Computational Linguistics: Human Language
  Technologies, Volume 1 (Long and Short Papers)}, pages 1073--1094.

\bibitem[{Liu et~al.(2019{\natexlab{b}})Liu, Ott, Goyal, Du, Joshi, Chen, Levy,
  Lewis, Zettlemoyer, and Stoyanov}]{liu2019roberta}
Yinhan Liu, Myle Ott, Naman Goyal, Jingfei Du, Mandar Joshi, Danqi Chen, Omer
  Levy, Mike Lewis, Luke Zettlemoyer, and Veselin Stoyanov. 2019{\natexlab{b}}.
\newblock Roberta: A robustly optimized bert pretraining approach.
\newblock \emph{arXiv preprint arXiv:1907.11692}.

\bibitem[{McCoy et~al.(2019)McCoy, Pavlick, and Linzen}]{mccoy2019right}
Tom McCoy, Ellie Pavlick, and Tal Linzen. 2019.
\newblock Right for the wrong reasons: Diagnosing syntactic heuristics in
  natural language inference.
\newblock In \emph{Proceedings of the 57th Annual Meeting of the Association
  for Computational Linguistics}, pages 3428--3448.

\bibitem[{Mitchell and Lapata(2008)}]{mitchell2008vector}
Jeff Mitchell and Mirella Lapata. 2008.
\newblock Vector-based models of semantic composition.
\newblock In \emph{proceedings of ACL-08: HLT}, pages 236--244.

\bibitem[{Mitchell and Lapata(2010)}]{mitchell2010composition}
Jeff Mitchell and Mirella Lapata. 2010.
\newblock Composition in distributional models of semantics.
\newblock \emph{Cognitive science}, 34(8):1388--1429.

\bibitem[{Murdoch et~al.(2018)Murdoch, Liu, and Yu}]{murdoch2018beyond}
W~James Murdoch, Peter~J Liu, and Bin Yu. 2018.
\newblock Beyond word importance: Contextual decomposition to extract
  interactions from lstms.
\newblock In \emph{International Conference on Learning Representations}.

\bibitem[{Nandakumar et~al.(2019)Nandakumar, Baldwin, and
  Salehi}]{nandakumar2019well}
Navnita Nandakumar, Timothy Baldwin, and Bahar Salehi. 2019.
\newblock How well do embedding models capture non-compositionality? a view
  from multiword expressions.
\newblock In \emph{Proceedings of the 3rd Workshop on Evaluating Vector Space
  Representations for NLP}, pages 27--34.

\bibitem[{Paszke et~al.(2019)Paszke, Gross, Massa, Lerer, Bradbury, Chanan,
  Killeen, Lin, Gimelshein, Antiga et~al.}]{paszke2019pytorch}
Adam Paszke, Sam Gross, Francisco Massa, Adam Lerer, James Bradbury, Gregory
  Chanan, Trevor Killeen, Zeming Lin, Natalia Gimelshein, Luca Antiga, et~al.
  2019.
\newblock Pytorch: An imperative style, high-performance deep learning library.
\newblock In \emph{Advances in Neural Information Processing Systems}, pages
  8024--8035.

\bibitem[{Pavlick et~al.(2015)Pavlick, Rastogi, Ganitkevitch, Van~Durme, and
  Callison-Burch}]{pavlick2015ppdb}
Ellie Pavlick, Pushpendre Rastogi, Juri Ganitkevitch, Benjamin Van~Durme, and
  Chris Callison-Burch. 2015.
\newblock Ppdb 2.0: Better paraphrase ranking, fine-grained entailment
  relations, word embeddings, and style classification.
\newblock In \emph{Proceedings of the 53rd Annual Meeting of the Association
  for Computational Linguistics and the 7th International Joint Conference on
  Natural Language Processing (Volume 2: Short Papers)}, pages 425--430.

\bibitem[{Peters et~al.(2018)Peters, Neumann, Zettlemoyer, and
  Yih}]{peters2018dissecting}
Matthew Peters, Mark Neumann, Luke Zettlemoyer, and Wen-tau Yih. 2018.
\newblock Dissecting contextual word embeddings: Architecture and
  representation.
\newblock In \emph{Proceedings of the 2018 Conference on Empirical Methods in
  Natural Language Processing}, pages 1499--1509.

\bibitem[{Radford et~al.(2018)Radford, Narasimhan, Salimans, and
  Sutskever}]{radford2018improving}
Alec Radford, Karthik Narasimhan, Tim Salimans, and Ilya Sutskever. 2018.
\newblock Improving language understanding by generative pre-training.
\newblock \emph{URL https://s3-us-west-2. amazonaws.
  com/openai-assets/researchcovers/languageunsupervised/language understanding
  paper. pdf}.

\bibitem[{Radford et~al.(2019)Radford, Wu, Child, Luan, Amodei, and
  Sutskever}]{radford2019language}
Alec Radford, Jeffrey Wu, Rewon Child, David Luan, Dario Amodei, and Ilya
  Sutskever. 2019.
\newblock Language models are unsupervised multitask learners.
\newblock \emph{OpenAI Blog}, 1(8):9.

\bibitem[{Raffel et~al.(2020)Raffel, Shazeer, Roberts, Lee, Narang, Matena,
  Zhou, Li, and Liu}]{raffel2020exploring}
Colin Raffel, Noam Shazeer, Adam Roberts, Katherine Lee, Sharan Narang, Michael
  Matena, Yanqi Zhou, Wei Li, and Peter~J Liu. 2020.
\newblock Exploring the limits of transfer learning with a unified text-to-text
  transformer.
\newblock \emph{Journal of Machine Learning Research}, 21(140):1--67.

\bibitem[{Roberts et~al.(2020)Roberts, Raffel, and Shazeer}]{roberts2020much}
Adam Roberts, Colin Raffel, and Noam Shazeer. 2020.
\newblock How much knowledge can you pack into the parameters of a language
  model?
\newblock \emph{arXiv preprint arXiv:2002.08910}.

\bibitem[{Sanh et~al.(2019)Sanh, Debut, Chaumond, and
  Wolf}]{sanh2019distilbert}
Victor Sanh, Lysandre Debut, Julien Chaumond, and Thomas Wolf. 2019.
\newblock Distilbert, a distilled version of bert: smaller, faster, cheaper and
  lighter.
\newblock \emph{arXiv preprint arXiv:1910.01108}.

\bibitem[{Saphra and Lopez(2020)}]{saphra2020lstms}
Naomi Saphra and Adam Lopez. 2020.
\newblock \href {http://arxiv.org/abs/2010.04650} {Lstms compose (and learn)
  bottom-up}.

\bibitem[{Shwartz and Dagan(2019)}]{shwartz2019still}
Vered Shwartz and Ido Dagan. 2019.
\newblock Still a pain in the neck: Evaluating text representations on lexical
  composition.
\newblock \emph{Transactions of the Association for Computational Linguistics},
  7:403--419.

\bibitem[{Socher et~al.(2013)Socher, Perelygin, Wu, Chuang, Manning, Ng, and
  Potts}]{socher2013recursive}
Richard Socher, Alex Perelygin, Jean Wu, Jason Chuang, Christopher~D Manning,
  Andrew~Y Ng, and Christopher Potts. 2013.
\newblock Recursive deep models for semantic compositionality over a sentiment
  treebank.
\newblock In \emph{Proceedings of the 2013 conference on empirical methods in
  natural language processing}, pages 1631--1642.

\bibitem[{Tenney et~al.(2019{\natexlab{a}})Tenney, Das, and
  Pavlick}]{tenney2019bert}
Ian Tenney, Dipanjan Das, and Ellie Pavlick. 2019{\natexlab{a}}.
\newblock Bert rediscovers the classical nlp pipeline.
\newblock In \emph{Proceedings of the 57th Annual Meeting of the Association
  for Computational Linguistics}, pages 4593--4601.

\bibitem[{Tenney et~al.(2019{\natexlab{b}})Tenney, Xia, Chen, Wang, Poliak,
  McCoy, Kim, Van~Durme, Bowman, Das et~al.}]{tenney2019you}
Ian Tenney, Patrick Xia, Berlin Chen, Alex Wang, Adam Poliak, R~Thomas McCoy,
  Najoung Kim, Benjamin Van~Durme, Samuel Bowman, Dipanjan Das, et~al.
  2019{\natexlab{b}}.
\newblock What do you learn from context? probing for sentence structure in
  contextualized word representations.
\newblock In \emph{7th International Conference on Learning Representations,
  ICLR 2019}.

\bibitem[{Vaswani et~al.(2017)Vaswani, Shazeer, Parmar, Uszkoreit, Jones,
  Gomez, Kaiser, and Polosukhin}]{vaswani2017attention}
Ashish Vaswani, Noam Shazeer, Niki Parmar, Jakob Uszkoreit, Llion Jones,
  Aidan~N Gomez, {\L}ukasz Kaiser, and Illia Polosukhin. 2017.
\newblock Attention is all you need.
\newblock In \emph{Advances in neural information processing systems}, pages
  5998--6008.

\bibitem[{Vig and Belinkov(2019)}]{vig2019analyzing}
Jesse Vig and Yonatan Belinkov. 2019.
\newblock Analyzing the structure of attention in a transformer language model.
\newblock In \emph{Proceedings of the 2019 ACL Workshop BlackboxNLP: Analyzing
  and Interpreting Neural Networks for NLP}, pages 63--76.

\bibitem[{Wang et~al.(2018)Wang, Singh, Michael, Hill, Levy, and
  Bowman}]{wang2018glue}
Alex Wang, Amanpreet Singh, Julian Michael, Felix Hill, Omer Levy, and Samuel
  Bowman. 2018.
\newblock Glue: A multi-task benchmark and analysis platform for natural
  language understanding.
\newblock In \emph{Proceedings of the 2018 EMNLP Workshop BlackboxNLP:
  Analyzing and Interpreting Neural Networks for NLP}, pages 353--355.

\bibitem[{Wieting et~al.(2015)Wieting, Bansal, Gimpel, and
  Livescu}]{wieting2015towards}
John Wieting, Mohit Bansal, Kevin Gimpel, and Karen Livescu. 2015.
\newblock Towards universal paraphrastic sentence embeddings.
\newblock \emph{arXiv preprint arXiv:1511.08198}.

\bibitem[{Wilcox et~al.(2018)Wilcox, Levy, Morita, and Futrell}]{wilcox2018rnn}
Ethan Wilcox, Roger Levy, Takashi Morita, and Richard Futrell. 2018.
\newblock What do {RNN} language models learn about filler--gap dependencies?
\newblock In \emph{Proceedings of the 2018 EMNLP Workshop BlackboxNLP:
  Analyzing and Interpreting Neural Networks for NLP}, pages 211--221.

\bibitem[{Wolf et~al.(2019)Wolf, Debut, Sanh, Chaumond, Delangue, Moi, Cistac,
  Rault, Louf, Funtowicz, and Brew}]{Wolf2019HuggingFacesTS}
Thomas Wolf, Lysandre Debut, Victor Sanh, Julien Chaumond, Clement Delangue,
  Anthony Moi, Pierric Cistac, Tim Rault, R'emi Louf, Morgan Funtowicz, and
  Jamie Brew. 2019.
\newblock Huggingface's transformers: State-of-the-art natural language
  processing.
\newblock \emph{ArXiv}, abs/1910.03771.

\bibitem[{Yang et~al.(2019{\natexlab{a}})Yang, Zhang, Tar, and
  Baldridge}]{pawsx2019emnlp}
Yinfei Yang, Yuan Zhang, Chris Tar, and Jason Baldridge. 2019{\natexlab{a}}.
\newblock {PAWS-X: A Cross-lingual Adversarial Dataset for Paraphrase
  Identification}.
\newblock In \emph{Proc. of EMNLP}.

\bibitem[{Yang et~al.(2019{\natexlab{b}})Yang, Dai, Yang, Carbonell,
  Salakhutdinov, and Le}]{yang2019xlnet}
Zhilin Yang, Zihang Dai, Yiming Yang, Jaime Carbonell, Russ~R Salakhutdinov,
  and Quoc~V Le. 2019{\natexlab{b}}.
\newblock Xlnet: Generalized autoregressive pretraining for language
  understanding.
\newblock In \emph{Advances in neural information processing systems}, pages
  5754--5764.

\bibitem[{Yin et~al.(2020)Yin, Meng, and Chang}]{yin2020sentibert}
Da~Yin, Tao Meng, and Kai-Wei Chang. 2020.
\newblock \href {http://arxiv.org/abs/2005.04114} {Sentibert: A transferable
  transformer-based architecture for compositional sentiment semantics}.

\bibitem[{Zhang et~al.(2019)Zhang, Baldridge, and He}]{paws2019naacl}
Yuan Zhang, Jason Baldridge, and Luheng He. 2019.
\newblock {PAWS: Paraphrase Adversaries from Word Scrambling}.
\newblock In \emph{Proc. of NAACL}.

\end{thebibliography}
\bibliographystyle{acl_natbib}

\end{document}